\documentclass[journal]{IEEEtran}

\ifCLASSINFOpdf
\else
   \usepackage[dvips]{graphicx}
\fi
\usepackage{url}

\usepackage{graphicx}
\usepackage{multirow}
\usepackage{lipsum}
\usepackage{mathtools}
\usepackage{cuted}
\usepackage{filecontents}
\usepackage[noadjust]{cite}

\makeatletter
\def\@citex[#1]#2{\leavevmode
\let\@citea\@empty
\@cite{\@for\@citeb:=#2\do
{\@citea\def\@citea{ -\penalty\@m\ }%
\edef\@citeb{\expandafter\@firstofone\@citeb\@empty}%
\if@filesw\immediate\write\@auxout{\string\citation{\@citeb}}\fi
\@ifundefined{b@\@citeb}{\hbox{\reset@font\bfseries ?}%
\G@refundefinedtrue
\@latex@warning
{Citation `\@citeb' on page \thepage \space undefined}}%
{\@cite@ofmt{\csname b@\@citeb\endcsname}}}}{#1}}
\makeatother

\begin{document}

\title{Semi-supervised and Unsupervised Methods for Heart Sounds Classification in Restricted Data Environments}

\author{Balagopal Unnikrishnan$^{1}$, Pranshu Ranjan Singh$^{1}$, Xulei Yang$^{2}$, and Matthew Chin Heng Chua$^{1}$
\thanks{1 Institute of Systems Science, National University of Singapore}
\thanks{2 Institute for Infocomm Research, A*STAR Singapore}
}

\maketitle

\begin{abstract}
Automated heart sounds classification is a much-required diagnostic tool in the view of increasing incidences of heart related diseases worldwide. In this study, we conduct a comprehensive study of heart sounds classification by using  various  supervised, semi-supervised and unsupervised approaches on the PhysioNet/CinC 2016 Challenge dataset. Supervised approaches, including deep learning and machine learning methods, require large amounts of labelled data to train the models, which are challenging to obtain in most practical scenarios. In view of the need to reduce the labelling burden for clinical practices, where human labelling is both expensive and time-consuming, semi-supervised or even unsupervised approaches in restricted data setting are desirable. A GAN based semi-supervised method is therefore proposed, which allows the usage of unlabelled data samples to boost the learning of data distribution. It achieves a better performance in terms of AUROC over the supervised baseline when limited data samples exist. Furthermore, several unsupervised methods are explored as an alternative approach by considering the given problem as an anomaly detection scenario. In particular, the unsupervised feature extraction using 1D CNN Autoencoder coupled with one-class SVM obtains good performance without any data labelling. The potential of the proposed semi-supervised and unsupervised methods may lead to a workflow tool in the future for the creation of higher quality datasets.
\end{abstract}


\IEEEpeerreviewmaketitle

\section{Introduction}
\IEEEPARstart{C}{ardiovascular} diseases (CVDs) have been the main cause of death globally. 17.9 million deaths have been attributed to CVDs, which represents 31\% of all global deaths \cite{ref1}. There is a need for methods for first hand examination of cardiovascular system. Auscultation of the heart sounds or Phonocardiogram (PCG) signals is a crucial component of physical examination and can help detect cardiac conditions such as arrhythmia, valve disease, heart failure, and more \cite{ref2}. Heart sound analysis by auscultation has been done by physicians to assess the heart condition over a period of time. However, designing an accurate and automated system for detection of abnormal heart sounds is challenging due to unavailability of rigorously validated and high-quality heart sounds datasets \cite{ref3}. 

Apart from PCG signals, Electrocardiogram (ECG) signals has been used for detecting arrhythmia, myocardial ischemia and chronic alterations \cite{ref4, ref5}. Although ECG signals can reveal various intricate and abnormal heart behaviors, symptoms such as heart murmurs are concealed from an ECG signal \cite{ref6}. The use of heart sounds to detect various heart abnormalities has led to the development of wide range of algorithms. In \cite{ref7}, PCG signals undergo digital subtraction analysis to detect and characterize heart murmurs. Automated classification methods of heart sounds involve approaches such as Support Vector Machines (SVM) \cite{ref8}, Neural Networks \cite{ref9}, Probability based methods \cite{ref10} and ensemble of various classifiers \cite{ref11}.

The design of supervised methods for heart sounds classification requires large amount of labelled data. However, it is often difficult, expensive, or time-consuming to obtain additional labelled data \cite{ref12}. There are challenges in obtaining patient’s data in the medical domain. Furthermore, multiple physicians have to perform labelling in order to achieve a common consensus, etc. Semi-supervised learning and active learning methods deal with this problem by utilizing available unlabelled data along with the labelled data to build better classifier models \cite{ref13}. Chamberlain, Daniel, et al. demonstrate automatic lung sounds classification using a semi-supervised deep learning algorithm \cite{ref14}. Transfer learning for supervised heart sounds classification and data augmentation for minority class (abnormal category) samples are some of the areas being explored to improve the performance over traditional supervised classification methods \cite{ref15, ref16}.    

In most cases, the abnormal samples are much lesser than normal samples. This leads to a class imbalance when performing classification tasks \cite{ref17}. It is both time-consuming and expensive to collect the abnormal samples.There have been works that perform clustering on the extracted features from the heart sounds, followed by classification \cite{ref18}. In anomaly detection methods, the model is trained only on normal samples, but tested with both normal and abnormal samples \cite{ref43}.

In this work, the focus is on exploring current and new supervised, semi-supervised and unsupervised methods for heart sounds classification. The main contributions of this work are:\newline(i) Analysis of the performance of various supervised methods for heart sounds classification; \newline (ii) Utilization of  the Generative Adversarial Network (GAN) -based semi-supervised technique to obtain better performance in terms of Area Under the Receiver Operating Characteristic curve (AUROC) as compared to the supervised benchmark, and \newline (iii) Learning of latent representations from features of heart sounds using a 1D Convolutional Neural Network (CNN) model (Unsupervised method) and anomaly detection algorithms, and evaluate the classification performance using AUROC metric.\newline The methods and experimental analysis are discussed in detail in the following sections.  

\section{Methodology}

This section describes the data and the methods used in this study. The sub-section Dataset and Data Preparation describe the dataset used and the feature extraction methods for heart sounds, respectively. Subsequent sub-sections explain the techniques used for heart sound classification using supervised, semi-supervised and unsupervised methods.

\subsection{Dataset}

\begin{figure}
\centerline{\includegraphics[scale=0.5]{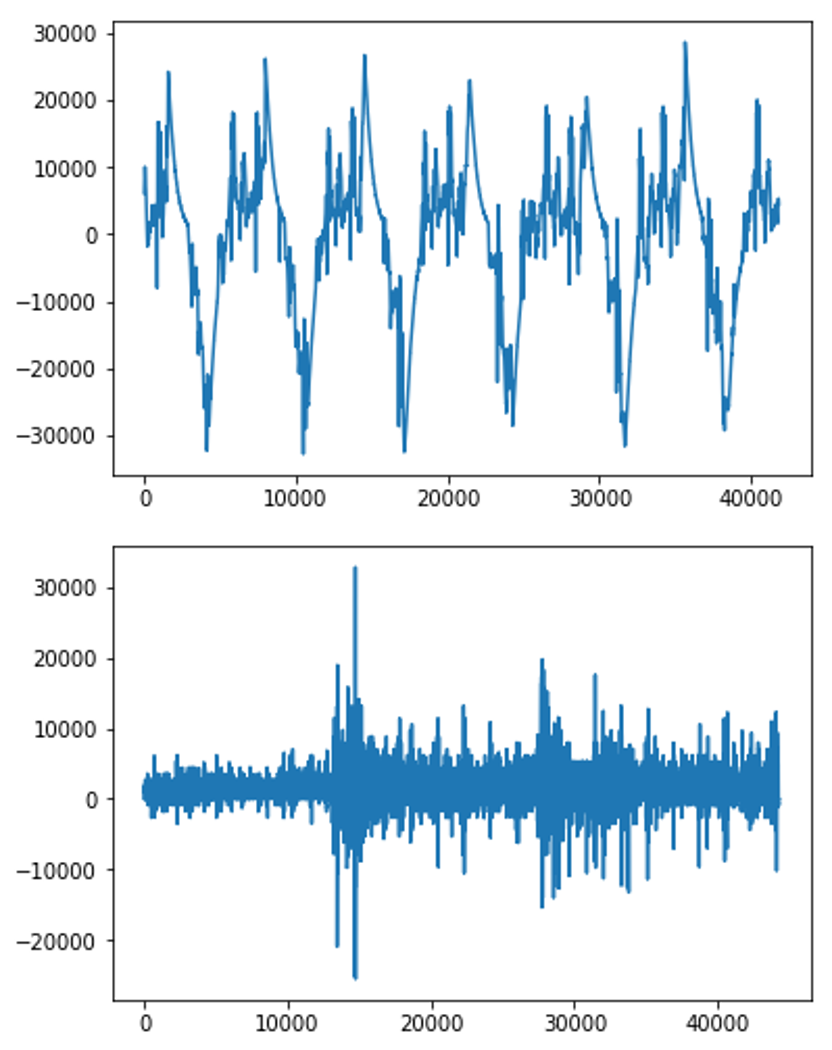}}
\caption{The heart sounds signal for normal class (top) and abnormal class (bottom). The x-axis represents the time-steps and y-axis represents the signal value. The sampling rate of the signal is 2000 Hz.}
\label{figure::sample_images}
\end{figure}

\begin{figure}
\centerline{\includegraphics[width=\columnwidth]{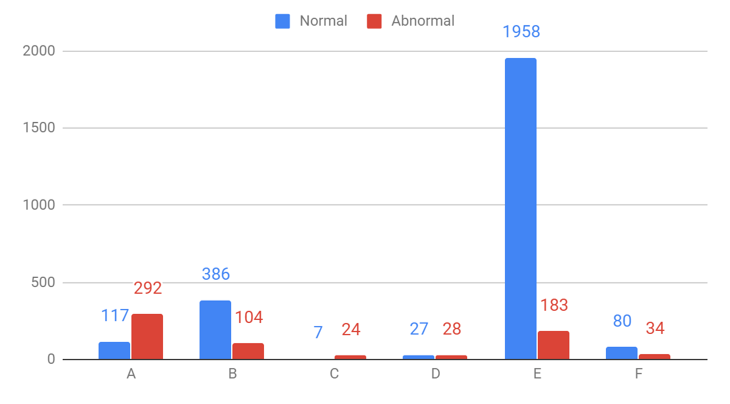}}
\caption{The 2016 PhysioNet/Computing in Cardiology Challenge dataset distribution. The dataset was obtained by combining heart sounds databases collected independently by various research teams. The individual datasets are labelled A, B, C, D, E and F. The distribution of normal and abnormal samples in each sub-dataset is different.}
\label{figure::data_distrbution}
\end{figure}

The heart sounds dataset used for this study was provided by the 2016 PhysioNet/Computing in Cardiology Challenge \cite{ref2}. It contains 3,240 labelled heart sounds recordings. The dataset is divided into two classes, Normal and Abnormal samples. Fig. \ref{figure::sample_images} shows the heart sounds signal for normal and abnormal sample. The duration of heart sounds signal ranges from 5 seconds (short-period) to 120 seconds (long period). This dataset was obtained by combining various heart sounds databases. It consists of 6 sub-datasets labelled A, B, C, D, E and F as shown in the Fig. \ref{figure::data_distrbution}.

The heart sounds recordings in this dataset were collected from nine different locations of the body. The four major locations are the aortic area, pulmonic area, tricuspid area and mitral area.The normal recordings correspond to healthy subjects whereas the abnormal ones were obtained from patients with confirmed cardiac diagnosis. The typical illnesses of the patients were heart valve defects and coronary artery. The presence of noise in some samples were due to the uncontrolled environment of the recordings. The noise sources includes talking, stethoscope motion, breathing and intestinal sounds. 

\subsection{Data Preparation}

For this study, various features obtained from heart sounds signals are used for training different models. The raw signal undergoes pre-processing steps such as padding and pruning. For padding operation, all the samples are zero-padded to achieve the length of the maximum length signal (120 seconds) in the dataset. For pruning operation, all the signals are truncated to achieve the length of minimum length signal (5 seconds) in the dataset.  

\begin{figure}
\centerline{\includegraphics[width=\columnwidth]{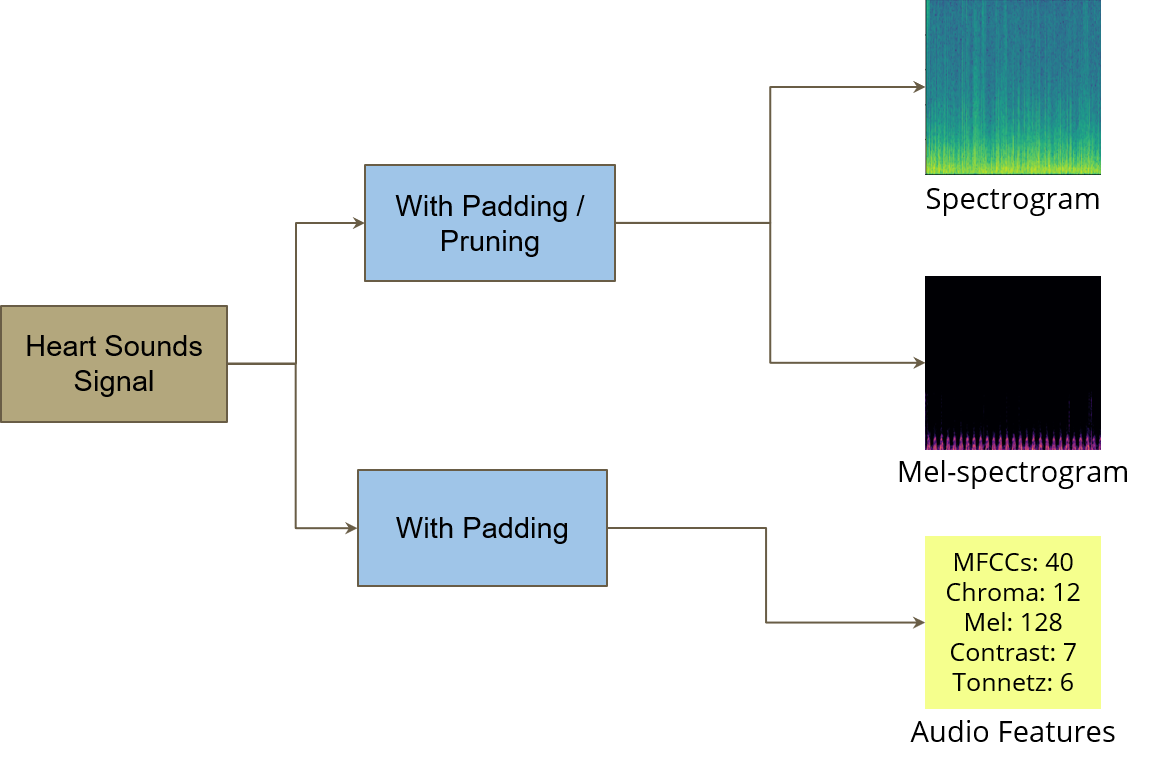}}
\caption{Feature Extraction from Heart Sounds Signal. Various features are extracted for supporting various techniques of heart sounds classification. spectrogram and Mel-spectrogram are obtained by converting the PCG signals to image. Audio features are obtained by appending specific features such as MFCC sequence, Chroma, Mel-spectrogram, Contrast and Tonnetz.}
\label{figure::feature_extraction}
\end{figure}

The different types of features extracted from the heart sounds signal are shown in Fig. \ref{figure::feature_extraction}. For semi-supervised methods, the raw processed signal is used as input. For supervised methods, both the padded and pruned signals are used to obtain the spectrogram and mel-spectrogram features. Both spectrogram and mel-spectrogram features are plotted with the time as the x-axis and frequency as y-axis. These plots are saved in form of color images having resolution 64 x 64 x 3 and 128 x 128 x 3 respectively. 

Audio features such as Mel-Frequency Cepstral Coefficients (MFCCs), Chroma \cite{ref19}, mel-scaled spectrogram (mel-spectrogram), spectral contrast \cite{ref20} and tonal centroid features (tonnetz) \cite{ref21} were extracted from the heart sounds signals. MFCCs, Chroma, mel-spectrogram, Spectral Contrast and Tonnetz contribute 40, 12, 128, 7 and 6 features, respectively. These features are appended to form a combined feature list with 193 features. These extracted audio features are used in supervised methods and unsupervised methods (for anomaly detection). Since there is a class imbalance, oversampling of minority class (Abnormal class) is performed using Synthetic Minority Over-sampling Technique (SMOTE) \cite{ref22}. This oversampling is performed on the audio features.  

\subsection{Supervised Methods for Heart Sounds 
Classification}

The various supervised methods used for performing heart sounds classification can be grouped in four clusters:\newline(i) Transfer Learning using pre-trained deep learning models on spectrogram/ Mel-spectrogram images;\newline(ii) Custom CNN on spectrogram images;\newline(iii) Deep Learning models on extracted audio features, and (iv) Machine Learning models on extracted audio features.\newline
The details of the methods are described below.

\subsubsection{Transfer Learning using Pre-trained Deep Learning Models on Spectrogram/ Mel-spectrogram Images}

Transfer learning in CNNs has shown that the image representations learnt over a large-scale labelled dataset can be transferred to classification tasks over limited data samples \cite{ref23}. ResNet-50 \cite{ref24}, Inception-v3 \cite{ref25} and DenseNet-121 \cite{ref26} have shown state-of-the-art classification results on the ImageNet dataset. The spectrograms and mel-spectrograms obtained from heart sounds signals are converted to 64 x 64 x 3 images. (from Data Preparation sub-section) These images are trained on ImageNet pre-trained ResNet-50, Inception-v3 and DenseNet-121 models. The output of the final convolutional layer of three models is fed to a fully-connected single node layer for classification into Normal or Abnormal class.       

\subsubsection{Custom CNN on Spectrogram Images}

The spectrogram obtained from the heart sounds signal is converted to 128 x 128 x 3 image. These images are fed to a custom designed CNN network which follows VGG \cite{ref27} like architecture. The custom architecture of Custom CNN is provided in the Table \ref{table::tab1}. The input spectrogram image passes through a series of convolution and pooling layers, and dense layers towards the end of the network to output the class of the heart sounds signal. ReLU activation \cite{ref37} has been used for the convolutional and dense layers, except for the final dense layer, which uses Sigmoid activation.  Dropout layers are added to prevent the model from over-fitting to the training set \cite{ref36}.       

\begin{table}
\caption{Custom CNN Architecture on Spectrogram Images}
\label{table::tab1}
\small
\setlength{\tabcolsep}{3pt}
\begin{tabular}{|p{80pt}|p{160pt}|}
\hline
Layers & Attributes \\ \hline
Convolution 2D & 16 filters, 3 x 3 kernel, ReLU activation, padding=same \\ \hline
Convolution 2D & 16 filters, 3 x 3 kernel, ReLU activation, padding=same \\ \hline
MaxPool 2D & 2 x 2 kernel, stride=2 \\ \hline
Convolution 2D & 32 filters, 3 x 3 kernel, ReLU activation, padding=same \\ \hline
Convolution 2D & 32 filters, 3 x 3 kernel, ReLU activation, padding=same \\ \hline
MaxPool 2D & 2 x 2 kernel, stride=2 \\ \hline
Convolution 2D & 64 filters, 3 x 3 kernel, ReLU activation, padding=same \\ \hline
Convolution 2D & 64 filters, 3 x 3 kernel, ReLU activation, padding=same \\ \hline
MaxPool 2D & 2 x 2 kernel, stride=2 \\ \hline
Convolution 2D & 128 filters, 3 x 3 kernel, ReLU activation, padding=same \\ \hline
Convolution 2D & 128 filters, 3 x 3 kernel, ReLU activation, padding=same \\ \hline
MaxPool 2D & 2 x 2 kernel, stride=2 \\ \hline
Flatten \& Dropout & dropout rate=0.25 \\ \hline
Dense & 256 nodes, ReLU activation \\ \hline
Dropout & dropout rate=0.25 \\ \hline
Dense & 1 node, Sigmoid activation \\ \hline
\end{tabular}
\end{table}

\subsubsection{Deep Learning Models on Extracted Audio Features}

The audio features extracted from the heart sounds signals undergo oversampling using SMOTE to obtain the equal number of samples for both Normal and Abnormal classes. These features are then modeled using Dense Neural Network (Dense NN), Neural Network with Long Short Term Memory units (LSTM NN) \cite{ref28} and 1D CNN. The Dense NN architecture takes a feature list of dimension 193 as input as passes it through a series of densely connected layers. The Dense NN architecture consists of 4 dense layers, each having 128 nodes with ReLU activation, followed by a single node densely connected layer with Sigmoid activation. The LSTM NN architecture takes a feature list of dimension 193 x 1 as input and passes it through a series of LSTM units and densely connected layers. The architecture for LSTM NN is provided in Table \ref{table::tab2}. The LSTM units are useful in modeling sequential data and final densely connected node is used for the classification. The 1D CNN architecture takes a feature list of dimension 193 x 1 as input and passes it through a series of 1D convolutional, 1D pooling and densely connected layers as depicted in Table \ref{table::tab3}.        

\begin{table}
\caption{Neural Network with LSTM units on Extracted Audio Features}
\label{table::tab2}
\small
\setlength{\tabcolsep}{3pt}
\begin{tabular}{|p{80pt}|p{160pt}|}
\hline
Layers & Attributes \\ \hline
LSTM & 128 units, dropout=0.2, recurrent dropout=0.25 \\ \hline
Dropout & dropout rate=0.25 \\ \hline
LSTM & 64 units, dropout=0.2, recurrent dropout=0.25 \\ \hline
Dense & 1 node, Sigmoid activation \\ \hline
\end{tabular}
\end{table}

\begin{table}
\caption{1D CNN on Extracted Audio Features}
\label{table::tab3}
\small
\setlength{\tabcolsep}{3pt}
\begin{tabular}{|p{80pt}|p{160pt}|}
\hline
Layers & Attributes \\ \hline
Convolution 1D & 128 filters, kernel size=3, ReLU activation \\ \hline
Convolution 1D & 128 filters, kernel size=3, ReLU activation \\ \hline
MaxPool 1D & kernel size=3, stride=3 \\ \hline
Convolution 1D & 256 filters, kernel size=3, ReLU activation \\ \hline
Convolution 1D & 256 filters, kernel size=3, ReLU activation \\ \hline
MaxPool 1D & kernel size=3, stride=3 \\ \hline
Convolution 1D & 512 filters, kernel size=3, ReLU activation \\ \hline
Convolution 1D & 512 filters, kernel size=3, ReLU activation \\ \hline
Flatten & $\_$ \\ \hline
Dense & 256 nodes, ReLU activation \\ \hline
Dense & 128 nodes, ReLU activation \\ \hline
Dense & 1 node, Sigmoid activation \\ \hline
\end{tabular}
\end{table}

\subsubsection{Machine Learning Models on Extracted Audio Features}

The extracted audio features are used to fit machine learning models such as Decision Tree, SVM, Random Forest and Gradient Boosting. For each machine learning method, the model was fitted on training set and hyper-parameters of the models were tuned using a validation set and finally evaluated on a test set. The various hyper-parameters and their values for each machine learning model are provided in Table \ref{table::tab4}. For decision tree modeling, the hyper-parameters are Criterion (the function that measures the quality of the split), max depth (the maximum depth of the decision tree), max leaf nodes (maximum number of nodes allowed in leaf/terminal positions) and class weight. The class weight hyper-parameter represents the ratio (abnormal$:$normal ratio) of weights associated with the classes. The hyper-parameters used in SVM modeling are C (penalty factor of error term), kernel (the type of kernel used in the algorithm), gamma (kernel coefficient) and class weight. The kernel used in SVM model was radial basis function (rbf). The hyper-parameters used in Random Forest modeling are Criterion, Number of estimators (the number of trees in the forest), max depth and max leaf nodes. For gradient boosting modeling, the hyper-parameters used are Number of estimators (the number of boosting stages), max depth and learning rate (the reduces contribution of each tree by this rate). The above hyper-parameters are tuned for each machine learning model for two cases, without SMOTE balancing and with SMOTE balancing.        

\begin{table}
\caption{Hyper-parameters for various machine learning models on extracted audio features}
\label{table::tab4}
\small
\setlength{\tabcolsep}{3pt}
\begin{tabular}{|p{126pt}|p{70pt}|p{35pt}|}
\hline
ML Model & Hyper-parameter & Value \\ \hline
\multirow{4}{*}{Decision Tree} & Criterion & Entropy \\ 
{} & Max Depth & $40$ \\
{} & Max Leaf Nodes & $40$ \\ 
{} & Class weight & $5$$:$$1$\\ \hline
\multirow{3}{*}{Decision Tree (with SMOTE)} & Criterion & Entropy \\ 
{} & Max Depth & $60$ \\
{} & Max Leaf Nodes & $40$ \\ \hline
\multirow{4}{*}{SVM} & C & $0.07$ \\ 
{} & Kernel & rbf \\
{} & Gamma & auto \\ 
{} & Class weight & $19$$:$$3$\\ \hline
\multirow{3}{*}{SVM (with SMOTE)} & C & $70$ \\ 
{} & Kernel & rbf \\
{} & Gamma & auto \\ \hline
\multirow{5}{*}{Random Forest} & Criterion & Entropy \\ 
{} & No. of estimators & $400$ \\
{} & Max Depth & $10$ \\
{} & Max Leaf Nodes & $50$ \\ 
{} & Class weight & $5$$:$$1$\\ \hline
\multirow{4}{*}{Random Forest (with SMOTE)} & Criterion & Entropy \\ 
{} & No. of estimators & $100$ \\
{} & Max Depth & $10$ \\
{} & Max Leaf Nodes & $64$ \\ \hline
\multirow{3}{*}{Gradient Boosting} & No. of estimators & $400$ \\
{} & Max Depth & $7$ \\
{} & Learning rate & $0.1$ \\ \hline
\multirow{3}{*}{Gradient Boosting (with SMOTE)} & No. of estimators & $400$ \\
{} & Max Depth & $6$ \\
{} & Learning rate & $0.1$ \\ \hline
\end{tabular}
\end{table}

\subsection{Semi-supervised Method: Generative Adversarial Network}

\begin{figure*}
\centering
\includegraphics[scale=0.60]{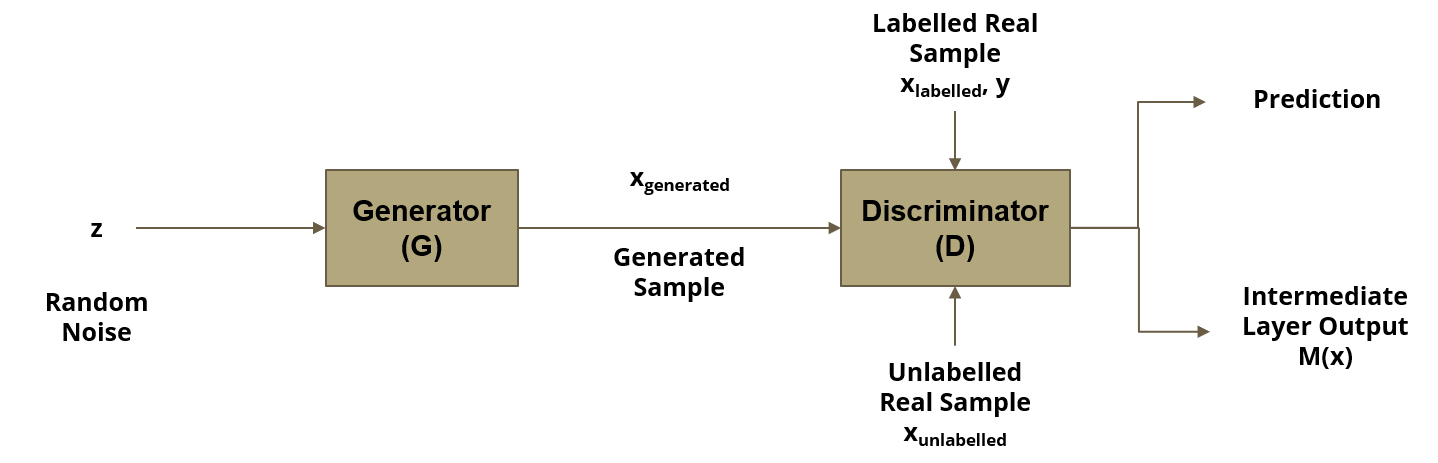}
\caption{Semi-supervised GAN Training Framework. The Generator (G) takes a random noise $z$ as input and produces a generated sample $x_{generated}$. The Discriminator (D) takes the generated samples, labelled real samples ($x_{labelled}$, $y$) and unlabelled real samples $x_{unlabelled}$ and produces the prediction of the class label and the Intermediate layer output $M(x)$.}
\label{figure::gan_training}
\end{figure*}

\begin{figure}
\centering
\includegraphics[width=4.5cm]{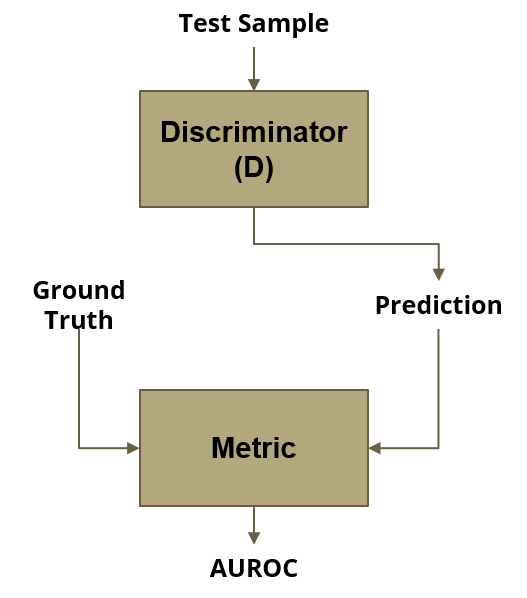}
\caption{Semi-supervised GAN Testing Framework. During the testing phase, only Discriminator (D) is used. The test sample is fed to the Discriminator to obtain the class prediction. The predicted class along with the ground truth class is used to obtain the AUROC metric.}
\label{figure::gan_testing}
\end{figure}

Generative adversarial networks (GANs) has provided a way for generating fake samples and utilizing them for other tasks \cite{ref32}. The semi-supervised method makes use of GANs to utilize the unlabelled data samples. The semi supervised models has access to both the labelled and unlabelled data from the training set. In theory, such models should perform better than the supervised methods as they now have access to unlabelled training data - provided the semi-supervised smoothness assumption holds i.e. if two points x1, x2 are close in a high density region, their labels y1, y2 are also be closer \cite{ref33}. This class of semi-supervised algorithms are called generative models and they are generally trained in a coupled fashion, similar to the training procedure of GANs. Fig. \ref{figure::gan_training} and Fig. \ref{figure::gan_testing} provide the GAN training and testing framework. A combined loss function as mentioned in equations (\ref{eq1}) to (\ref{eq4}) is used to train the discriminator and generator, and the reformulation trick is used as depicted in \cite{ref29}. 

The generator is trained by matching the features of the generated samples and the real samples. The supervised loss is similar to the cross-entropy loss in K-class classification problems and the unsupervised loss helps in distinguishing between real and fake samples. This coupled training in an adversarial setting is used to train the semi-supervised network. The network architecture for the discriminator and generator are provided in Table \ref{table::arch_disc} and Table \ref{table::arch_gen}. 1D convolutions are used in both discriminator and generator network architecture. These are highly effective as convolution operations are translation and scale invariant and can pickup relevant features anywhere within the input. This is useful since the heart sounds are not segmented or aligned in any fashion. The first 5 seconds of the heart sounds data is directly taken as input for the semi-supervised method. In the overall training setup, minimal amount of annotation or labelling is required.  

\begin{figure*}
\centering
\begin{equation}
Loss_{discriminator} = Loss_{unsupervised} +  Loss_{supervised}
\label{eq1}
\end{equation}
\begin{equation}
Loss_{supervised} = - E(x, y) [log P_D(y|x, y < K + 1)], K = no. of classes
\label{eq2}
\end{equation}
\begin{equation}
Loss_{unsupervised} = -E_x [log (1 - P_D(y = K + 1|x))] - E_{xg} [log P_D(y = K + 1|x)] 
\label{eq3}
\end{equation}
\begin{equation}
Loss_{generator} = || E_{x} [M(x)] - E_{xg} [M(G(z))] ||^2_{2} 
\label{eq4}
\end{equation}
\newline 
\end{figure*}





\begin{table}
\caption{Semi-supervised GAN Discriminator Architecture}
\label{table::arch_disc}
\small
\setlength{\tabcolsep}{3pt}
\begin{tabular}{|p{80pt}|p{160pt}|}
\hline
Layers & Attributes \\ \hline
Convolution 1D & 64 filters, kernel size=8, stride=1, Leaky ReLU activation \\ \hline
Convolution 1D & 64 filters, kernel size=8, stride=2, Leaky ReLU activation \\ \hline
Convolution 1D & 128 filters, kernel size=8, stride=2, Leaky ReLU activation \\ \hline
Convolution 1D & 256 filters, kernel size=8, stride=2, Leaky ReLU activation \\ \hline
Convolution 1D & 256 filters, kernel size=8, stride=2, Leaky ReLU activation \\ \hline
Convolution 1D & 256 filters, kernel size=8, stride=2, Leaky ReLU activation \\ \hline
Convolution 1D & 256 filters, kernel size=8, stride=2, Leaky ReLU activation \\ \hline
Convolution 1D & 256 filters, kernel size=8, stride=2, Leaky ReLU activation \\ \hline
Adaptive Avg Pooling 1D & output size=1 \\ \hline
Flatten & Intermediate Layer Output \\ \hline
Dense & 2 nodes (number of classes) \\ \hline
\end{tabular}
\end{table}

\begin{table}
\caption{Semi-supervised GAN Generator Architecture}
\label{table::arch_gen}
\small
\setlength{\tabcolsep}{3pt}
\begin{tabular}{|p{80pt}|p{160pt}|}
\hline
Layers & Attributes \\ \hline
Dense & 256*33(=8448) nodes, batch norm 1D, ReLU activation \\ \hline
Reshape & Reshape to 256 x 33 \\ \hline
Conv Transpose 1D & 256 filters, kernel size=8, stride=2, padding=0, batch norm 1D, ReLU activation \\ \hline
Conv Transpose 1D & 256 filters, kernel size=8, stride=2, padding=0, batch norm 1D, ReLU activation \\ \hline
Conv Transpose 1D & 256 filters, kernel size=8, stride=2, padding=0, batch norm 1D, ReLU activation \\ \hline
Conv Transpose 1D & 256 filters, kernel size=8, stride=2, padding=0, batch norm 1D, ReLU activation \\ \hline
Conv Transpose 1D & 256 filters, kernel size=8, stride=2, padding=0, batch norm 1D, ReLU activation \\ \hline
Conv Transpose 1D & 128 filters, kernel size=8, stride=2, padding=1, batch norm 1D, ReLU activation \\ \hline
Conv Transpose 1D & 64 filters, kernel size=8, stride=2, padding=1, batch norm 1D, ReLU activation \\ \hline
Conv Transpose 1D & 1 filter, kernel size=8, stride=1, padding=0, tanh activation \\ \hline
\end{tabular}
\end{table}

\subsection{Unsupervised Method: Anomaly Detection}

For the purpose of obtaining good performance in restricted data environments, the method of anomaly detection was explored. In anomaly detection scenario, the model is trained using just the normal class samples. Any abnormality or deviation from normality is considered as a disease case (abnormal class). This has two major advantages, (i) it can perform the entire training without the need of any labels (need for abnormal class samples), and (ii) it allows for an anomaly score which can be used to get the relative grade of the abnormal samples and can be utilized in various applications such as triaging. 

Two anomaly detection algorithms and two sets of features (these features are used to train the two algorithms) are considered for evaluation. The two anomaly detection algorithms used are One-Class SVM \cite{ref31} and Isolation Forest \cite{ref30}. In One-Class SVM algorithm, the normal samples are enclosed within a hyper-sphere or hyper-plane and everything outside this is considered as an anomalous sample. The distance from the separating plane decides the degree of abnormality. In Isolation forest, the samples are split randomly during training using isolation trees. The resulting average tree lengths for the tree forest is taken as a measure of the abnormality. Anomalous samples are more susceptible to isolation while splitting and hence have shorter average tree lengths. This can be used to distinguish between the anomalies and normal samples. 

\begin{figure*}
\centering
\includegraphics[scale=0.6]{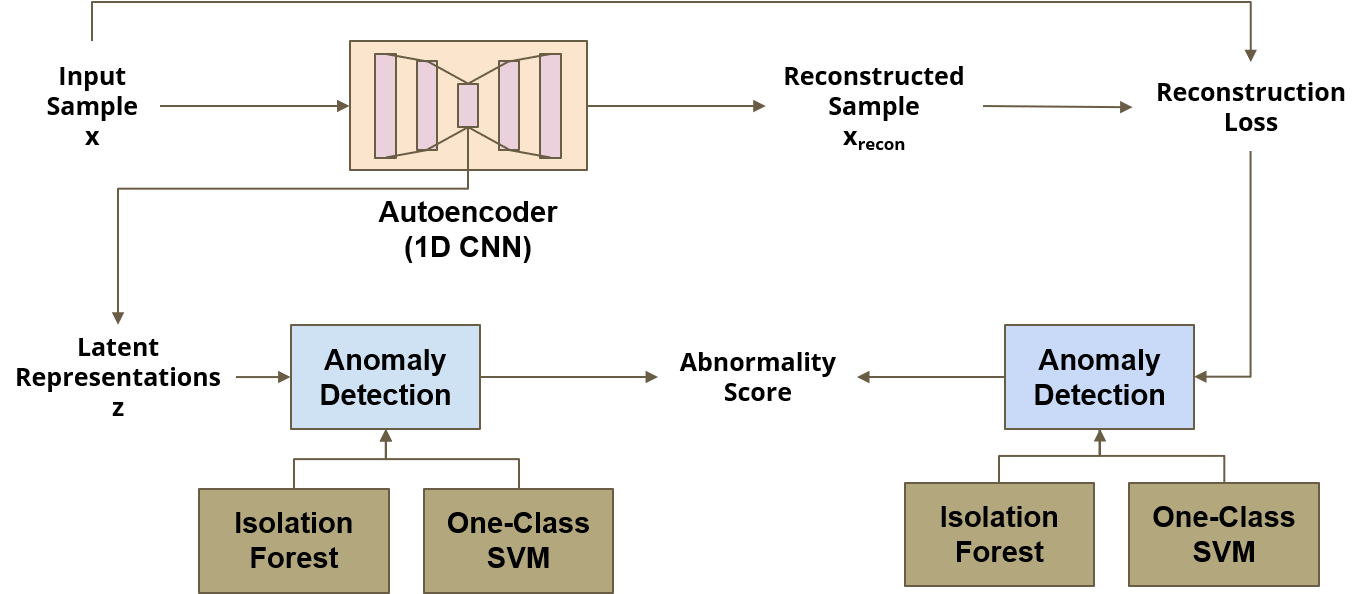}
\caption{Unsupervised Anomaly Detection Framework using 1D CNN Autoencoder. The Autoencoder takes an input sample $x$ and produces the reconstructed sample $x_{recon}$. The latent representations $z$ from the Autoencoder and the Reconstruction loss are used as features for anomaly detection methods, Isolation Forest and One-Class SVM.}
\label{figure::anomaly_detection}
\end{figure*}

During training, a stack of 1D Convolutions layers and 1D Convolutions-Upsampling layers are combined to serve as an auto-encoder. The audio features extracted from the heart sounds are provided to the auto-encoder for reconstruction. Two features from 1D CNN Autoencoder were used for serving as the input features for Isolation Forest and One-Class SVM:\newline
(i) Reconstruction loss: The squared difference between the actual input and the reconstructed output sample by the 1D CNN Autoencoder. The intuition is that, for the anomalous samples, the reconstruction loss would be higher during test time as the actual model cannot accurately reconstruct anomalous samples as it is trained to reconstruct only normal samples. The reconstruction loss is defined in equation \ref{eq5}. 
\begin{equation}
Loss_{reconstruction} = |X_{input}-X_{rec}|^2
\label{eq5}
\end{equation}
(ii) Latent Representations : Latent representations or embeddings is the output obtained from the bottleneck layer / the last layer of the encoder. While training, the latent representation would provide a set of features which represent the training samples. These feature set would help discriminate between normal and anomalous samples.

The overall anomaly detection framework is shown in Fig. \ref{figure::anomaly_detection}. The autoencoder network structure is provided in Table \ref{table::tab7}. Two modes of training were used for training the 1D CNN autoencoder. In the first case, the training data consists purely of normal samples only. In the second case, the data is contaminated with abnormal samples as well. This would help in evaluating the utility of the method in the use-case where there are no filters to prevent abnormal samples from being used - like screening applications - where the data can have a mix of both normal and abnormal samples, but the proportion of anomalous data is lesser. 8\% - 12\% is a reasonable assumption for contamination with anomalous samples as the percentage prevalence of heart diseases among general population is roughly 10\% \cite{ref34}. During the training phase, extracted latent representations and reconstruction loss for samples obtained from the auto-encoder are used to train the two anomaly detection algorithms. For both the algorithms, the experiments are conducted for clean data (only normal samples) and contaminated data (normal and abnormal mixed).

\begin{table}
\caption{1D CNN Autoencoder Architecture}
\label{table::tab7}
\small
\setlength{\tabcolsep}{3pt}
\begin{tabular}{|p{80pt}|p{160pt}|}
\hline
Layers & Attributes \\ \hline
Convolution 1D & 64 filters, kernel size=3, padding=same \\ \hline
MaxPool 1D & kernel size=2, stride=2 \\ \hline
Convolution 1D & 64 filters, kernel size=3, padding=same \\ \hline
MaxPool 1D & kernel size=2, stride=2 \\ \hline
Convolution 1D & 32 filters, kernel size=3, padding=same \\ \hline
MaxPool 1D & kernel size=2, stride=2 \\ \hline
Convolution 1D & 16 filters, kernel size=3, padding=same \\ \hline
MaxPool 1D & kernel size=2, stride=2 \\ \hline
Convolution 1D & 8 filters, kernel size=3, padding=same \\ \hline
MaxPool 1D & kernel size=2, stride=2 \\ \hline
Flatten & Latent Space \\ \hline
Reshape & Reshape to 12 x 8 \\ \hline
Convolution 1D & 8 filters, kernel size=3, padding=same \\ \hline
Upsampling 1D & size=2 \\ \hline
Convolution 1D & 16 filters, kernel size=3, padding=same \\ \hline
Upsampling 1D & size=2 \\ \hline
Convolution 1D & 32 filters, kernel size=3, padding=same \\ \hline
Upsampling 1D & size=2 \\ \hline
Convolution 1D & 64 filters, kernel size=3, padding=same \\ \hline
Upsampling 1D & size=2 \\ \hline
Zero Padding 1D & 0 x 1 \\ \hline
Convolution 1D & 1 filter, kernel size=3, padding=same \\ \hline
\end{tabular}
\end{table}

\section{Computations and Results}

This section describes the experiments performed to evaluate the methods discussed in previous section. Computational Setup and Evaluation Metrics sub-section describes the training setup and the various metrics used to validate the performance on a test set. Subsequent sub-section describes the results obtained for supervised, semi-supervised and unsupervised methods.   

\subsection{Computational Setup}

The computations of supervised methods for heart sounds classification utilize the entire dataset consisting of 3,240 samples. $20$\% of above dataset (648 samples) were used for testing and remaining $80$\% were used for training. This $80$\% was further divided into training ($90$\%, 2,333 samples) and validation ($10$\%, 259 samples) sets. The training set was used for model fitting and validation set was used to tune the hyper-parameters of the model.

The computations of semi-supervised and unsupervised methods utilize only the sub-dataset E (2141 samples) of the entire dataset. Since each of the sub-datasets are collected from different sources, there may be some bias associated with each sub-dataset. $20$\% of above sub-dataset E (429 samples) were used for testing and remaining $80$\% were used for training. This $80$\% was further divided into training ($90$\%, 1,540 samples) and validation ($10$\%, 172 samples) sets.

\subsection{Evaluation Metrics}

The supervised methods are evaluated using the standard classification metrics such as sensitivity, specificity and accuracy. An additional metric MAcc, which is defined as the average of sensitivity and specificity was also used for evaluation \cite{ref44}. For semi-supervised evaluation, the idea is to compare the performance of the supervised baseline against semi-supervised methods across different amounts of labelled data. The idea is to mimic a clinical use case setting where the number of labelled samples is limited. As there is class imbalance, AUROC (Area Under the Receiver Operating Characteristic curve) is used as the metric for comparison between supervised and semi-supervised models. AUROC not only takes into account the issue of class imbalance, but is also not sensitive to the cutoff value taken for the class predictions. 

\subsection{Results and Discussion}

Table \ref{table::tab9} shows the results for various supervised methods discussed in this study. DenseNet-121 on Mel-spectrograms (with padding) and Decision Tree on extracted audio features achieved the best performance in terms of specificity and sensitivity respectively. Gradient Boosting on extracted audio features (with SMOTE balancing) achieved the best performance in terms of accuracy and MAcc. Table \ref{table::tab_sup_results} shows the comparison among the Gradient Boosting method and the methods reported in the PhysioNet/CinC 2016 Challenge.

\begin{table}
\caption{Results for Supervised methods of Heart Sounds Classification}
\label{table::tab9}
\small
\setlength{\tabcolsep}{3pt}
\begin{tabular}{|p{80pt}|p{35pt}|p{40pt}|p{42pt}|p{25pt}|}
\hline
{Method} & {Accuracy} & {Specificity} & {Sensitivity} & {MAcc}\\ \hline
ResNet-50 on Mel-spectrogram (with padding) & $0.869$ & $0.941$ & $0.604$ & $0.773$  \\ \hline
ResNet-50 on spectrogram (with padding) & $0.878$ & $0.943$ & $0.640$ & $0.792$  \\ \hline
ResNet-50 on Mel-spectrogram (with pruning) & $0.860$ & $0.919$ & $0.640$ & $0.780$  \\ \hline
ResNet-50 on spectrogram (with pruning) & $0.847$ & $0.896$ & $0.669$ & $0.782$  \\ \hline
Inception-v3 on Mel-spectrogram (with padding) & $0.850$ & $0.931$ & $0.554$ & $0.743$  \\ \hline
Inception-v3 on spectrogram (with padding) & $0.867$ & $0.947$ & $0.576$ & $0.761$  \\ \hline
Inception-v3 on Mel-spectrogram (with pruning) & $0.796$ & $0.941$ & $0.266$ & $0.604$  \\ \hline
Inception-v3 on spectrogram (with pruning) & $0.826$ & $0.953$ & $0.360$ & $0.656$  \\ \hline
DenseNet-121 on Mel-spectrogram (with padding) & $0.869$ & $0.965$ & $0.518$ & $0.741$  \\ \hline
DenseNet-121 on spectrogram (with padding) & $0.818$ & $\textbf{0.990}$ & $0.187$ & $0.589$  \\ \hline
Custom CNN on spectrogram (with padding) & $0.909$ & $0.967$ & $0.698$ & $0.832$  \\ \hline
Dense NN on extracted audio features (with SMOTE) & $0.855$ & $0.880$ & $0.763$ & $0.821$  \\ \hline
LSTM NN on extracted audio features (with SMOTE) & $0.748$ & $0.770$ & $0.670$ & $0.720$  \\ \hline
1D CNN on extracted audio features (with SMOTE) & $0.843$ & $0.847$ & $0.827$ & $0.837$  \\ \hline
Decision Tree on extracted audio features & $0.824$ & $0.811$ & $\textbf{0.870}$ & $0.841$  \\ \hline
Decision Tree on extracted audio features (with SMOTE) & $0.832$ & $0.837$ & $0.813$ & $0.825$  \\ \hline
SVM on extracted audio features & $0.807$ & $0.813$ & $0.784$ & $0.799$  \\ \hline
SVM on extracted audio features (with SMOTE) & $0.827$ & $0.953$ & $0.367$ & $0.660$  \\ \hline
Random Forest on extracted audio features & $0.898$ & $0.925$ & $0.798$ & $0.862$  \\ \hline
Random Forest on extracted audio features (with SMOTE) & $0.878$ & $0.888$ & $0.842$ & $0.865$  \\ \hline
Gradient Boosting on extracted audio features & $0.913$ & $0.970$ & $0.705$ & $0.838$  \\ \hline
Gradient Boosting on extracted audio features (with SMOTE) & $\textbf{0.913}$ & $0.935$ & $0.834$ & $\textbf{0.885}$  \\ \hline
\end{tabular}
\end{table}

\begin{table}
\caption{Comparison of proposed method with various supervised methods reported in the PhysioNet/CinC 2016 Challenge}
\label{table::tab_sup_results}
\small
\setlength{\tabcolsep}{3pt}
\begin{tabular}{|p{100pt}|p{55pt}|p{40pt}|p{30pt}|}
\hline
{Method} & {Feature} & {Balancing data} & {MAcc} \\ \hline
AdaBoost and CNN \cite{ref11} & Time-frequency & No & $0.8602$ \\ \hline
Ensemble of NN \cite{ref38} & Time-frequency & Yes & $0.8590$ \\ \hline
Dropout Connected NN \cite{ref39} & MFCC & No & $0.8520$ \\ \hline
SVM and KNN \cite{ref40} & Time-frequency, MFCC & No & $0.8454$ \\ \hline
CNN \cite{ref41} & MFCC & No & $0.8399$ \\ \hline
SVM and ELM \cite{ref42} & Audio Signal Analysis & No & $0.7869$ \\ \hline
Gradient Boosting (Current study) & Extracted Audio features & Yes & $0.8850$ \\ \hline
\end{tabular}
\end{table}

\begin{figure}
\centerline{\includegraphics[width=\columnwidth]{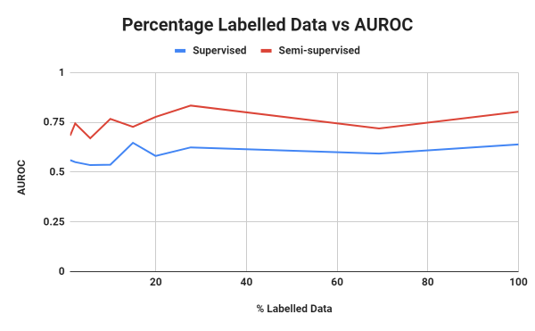}}
\caption{Semi-supervised Results. The graph shows the AUROC evaluation metric against the percentage of labelled data for supervised baseline and semi-supervised method. The performance of semi-supervised approach is better than supervised approach throughout the graph.}
\label{figure::semi-supervised_plot}
\end{figure}

For the semi-supervised computations, the percentage of labelled data provided to the models is slowly increased across the computation and compared based on the AUROC scores. Fig. \ref{figure::semi-supervised_plot} shows the plot of AUROC against the percentage of labelled data used for supervised and semi-supervised method. The performance for the semi-supervised model is better than the supervised baseline even in the case of very few data samples. Even with 4 or 8 data samples, the semi-supervised model is able to outperform the supervised baseline. This observation in the plot can be explained as: \newline (i) The larger unlabelled training set helps approximate the overall data distribution, which allows for a much better decision boundary than the supervised method which can only account for labelled samples; \newline (ii) The unlabelled data has a regularization effect on the classification network as the semi-supervised training follows a coupled adversarial training procedure. \newline A higher performance is obtained in terms of AUROC as more and more labelled samples are used for classification.

\textit{Use cases for semi-supervised methods:} These methods are of particular importance in cases of limited annotation ability. In most clinical cases, large number of labelled training samples for supervised training is not readily available. The semi-supervised methods can be used in two scenarios: \newline (i) It can learn better from lesser labelled data and provide better labels for pseudo-labelling algorithms; \newline (ii) It has a better predictive power as compared to supervised methods, and due to the usage of unlabelled samples as well, the model can iteratively select samples that needs to be labelled for better performance.

Table \ref{table::tab10} and Table \ref{table::tab11} provide the results of the anomaly detection methods. One-Class SVM achieves a better performance in terms of AUROC for both cases as compared to Isolation Forest. Moreover, latent representations or embeddings give a better performance than reconstruction loss. Data contamination in the experiment with autoencoder trained on only normal samples is not a major concern as the latent representations obtained are fairly robust to this issue. However, for the experiment with autoencoder trained on both normal and abnormal samples, it is ideal that normal samples be fed into the anomaly detection algorithm. 

\textit{Use cases for unsupervised methods:} The unsupervised feature extraction (using 1D CNN Autoencoder) method coupled with anomaly detection methods achieved good performances even with no major labelling burden. These methods can be used in two scenarios: \newline (i) These methods are good for triaging applications since the abnormality scores are an indicator of the disease and the ones with the higher abnormality score can be evaluated first; \newline (ii) These methods can be useful for creating datasets for supervised or semi-supervised training. Samples that are close to the classification boundary can be chosen in an unsupervised setting, as these samples are more confusing for the models to distinguish.

\begin{table}
\caption{Results for Anomaly Detection when autoencoder is trained on only normal samples}
\label{table::tab10}
\small
\setlength{\tabcolsep}{3pt}
\begin{tabular}{|p{78pt}|p{50pt}|p{60pt}|p{40pt}|}
\hline
{Method} & {Features} & {Labels} & {AUROC} \\ \hline
\multirow{4}{*}{Isolation Forest} & \multirow{2}{*}{Embeddings} & Normal & $0.644$ \\ {} & {} & Contaminated & $0.564$ \\ \cline{2-4} {} & \multirow{2}{*}{Rec Loss} & Normal & $0.699$ \\ {} & {} & Contaminated & $0.687$ \\ \hline  
\multirow{4}{*}{One-Class SVM} & \multirow{2}{*}{Embeddings} & Normal & $\textbf{0.842}$ \\ {} & {} & Contaminated & $0.552$ \\ \cline{2-4} {} & \multirow{2}{*}{Rec Loss} & Normal & $0.737$ \\ {} & {} & Contaminated & $0.658$ \\ \hline  
\end{tabular}
\end{table}

\begin{table}
\caption{Results for Anomaly Detection when autoencoder is trained on entire data (both normal and abnormal samples)}
\label{table::tab11}
\small
\setlength{\tabcolsep}{3pt}
\begin{tabular}{|p{78pt}|p{50pt}|p{60pt}|p{40pt}|}
\hline
{Method} & {Features} & {Labels} & {AUROC} \\ \hline
\multirow{4}{*}{Isolation Forest} & \multirow{2}{*}{Embeddings} & Normal & $0.678$ \\ {} & {} & Contaminated & $0.630$ \\ \cline{2-4} {} & \multirow{2}{*}{Rec Loss} & Normal & $0.577$ \\ {} & {} & Contaminated & $0.542$ \\ \hline  
\multirow{4}{*}{One-Class SVM} & \multirow{2}{*}{Embeddings} & Normal & $\textbf{0.828}$ \\ {} & {} & Contaminated & $0.567$ \\ \cline{2-4} {} & \multirow{2}{*}{Rec Loss} & Normal & $0.671$ \\ {} & {} & Contaminated & $0.588$ \\ \hline  
\end{tabular}
\end{table}

\section{Conclusion and Future Directions}

This study explores the supervised, semi-supervised and unsupervised methods of heart sounds classification for the use cases where the availability of labelled data is scarce. In such cases, the supervised methods with large number of labelled samples, plateau out and have similar performances. However, for smaller number of labelled samples, the semi-supervised algorithm outperforms the supervised baselines. Furthermore, the given problem is framed as an anomaly detection problem with unsupervised feature learning. The issue of data contamination is also studied and the results are presented. 

These works can be a starting point for various future use cases and studies. One promising direction for the utilization of these methods is in the case of active learning - where first a small subset of samples is labelled and then iteratively samples are chosen to be labelled further to improve performance. The good performance on lower number of labelled samples is also useful in the case for pseudo-labelling, where existing supervised classification methods are used with the assumed labels. The heart sounds signals used in this study are not segmented. Various segmentation algorithms have been developed in recent years. Proper segmentation and alignment techniques can be employed to further boost the performance. Apart from band-pass filters, other signal processing techniques can be explored to improve performance. Hence, better pre-processing techniques and feature extraction techniques can be another pathway for exploration. 

Another more challenging setting is to use data augmentation for heart sounds signals. It might be interesting to see how sound signals can be augmented using techniques like Mix-Up \cite{ref35}, apart from SMOTE for data balancing. However, since the domain of application is health care, it is important to ensure that the augmented data samples should not introduce any wrong features or biases within the model and hence should be undertaken with utmost care. Moreover, all the methods presented in this work can generalize for any 1D signal input. Hence, ECG signals can also be used instead of PCG signals. This work is presented with the belief that it can aid both in creation of better models and more importantly, better datasets which can further improve performance, as in most practical cases, it is the quality of data used that is a crucial factor in obtaining better performance. 

\bibliographystyle{IEEEtran}
\bibliography{heart_sound.bib}

\begin{thebibliography}{10}
\providecommand{\url}[1]{#1}
\csname url@samestyle\endcsname
\providecommand{\newblock}{\relax}
\providecommand{\bibinfo}[2]{#2}
\providecommand{\BIBentrySTDinterwordspacing}{\spaceskip=0pt\relax}
\providecommand{\BIBentryALTinterwordstretchfactor}{4}
\providecommand{\BIBentryALTinterwordspacing}{\spaceskip=\fontdimen2\font plus
\BIBentryALTinterwordstretchfactor\fontdimen3\font minus
  \fontdimen4\font\relax}
\providecommand{\BIBforeignlanguage}[2]{{%
\expandafter\ifx\csname l@#1\endcsname\relax
\typeout{** WARNING: IEEEtran.bst: No hyphenation pattern has been}%
\typeout{** loaded for the language `#1'. Using the pattern for}%
\typeout{** the default language instead.}%
\else
\language=\csname l@#1\endcsname
\fi
#2}}
\providecommand{\BIBdecl}{\relax}
\BIBdecl

\bibitem{ref1}
\BIBentryALTinterwordspacing
``Cardiovascular diseases (cvds),'' May 2017. [Online]. Available:
  \url{www.who.int/news-room/fact-sheets/detail/cardiovascular-diseases-(cvds)}
\BIBentrySTDinterwordspacing

\bibitem{ref2}
C.~Liu, D.~Springer, Q.~Li, B.~Moody, R.~A. Juan, F.~J. Chorro, F.~Castells,
  J.~M. Roig, I.~Silva, A.~E. Johnson \emph{et~al.}, ``An open access database
  for the evaluation of heart sound algorithms,'' \emph{Physiological
  Measurement}, vol.~37, no.~12, p. 2181, 2016.

\bibitem{ref3}
R.~M. Rangayyan and R.~J. Lehner, ``Phonocardiogram signal analysis: a
  review.'' \emph{Critical reviews in biomedical engineering}, vol.~15, no.~3,
  pp. 211--236, 1987.

\bibitem{ref4}
N.~V. Thakor and Y.-S. Zhu, ``Applications of adaptive filtering to ecg
  analysis: noise cancellation and arrhythmia detection,'' \emph{IEEE
  transactions on biomedical engineering}, vol.~38, no.~8, pp. 785--794, 1991.

\bibitem{ref5}
R.~Silipo and C.~Marchesi, ``Artificial neural networks for automatic ecg
  analysis,'' \emph{IEEE transactions on signal processing}, vol.~46, no.~5,
  pp. 1417--1425, 1998.

\bibitem{ref6}
W.~Phanphaisarn, A.~Roeksabutr, P.~Wardkein, J.~Koseeyaporn, and P.~Yupapin,
  ``Heart detection and diagnosis based on ecg and epcg relationships,''
  \emph{Medical devices (Auckland, NZ)}, vol.~4, p. 133, 2011.

\bibitem{ref7}
M.~A. Akbari, K.~Hassani, J.~D. Doyle, M.~Navidbakhsh, M.~Sangargir,
  K.~Bajelani, and Z.~S. Ahmadi, ``Digital subtraction phonocardiography (dsp)
  applied to the detection and characterization of heart murmurs,''
  \emph{Biomedical engineering online}, vol.~10, no.~1, p. 109, 2011.

\bibitem{ref8}
S.~Ari, K.~Hembram, and G.~Saha, ``Detection of cardiac abnormality from pcg
  signal using lms based least square svm classifier,'' \emph{Expert Systems
  with Applications}, vol.~37, no.~12, pp. 8019--8026, 2010.

\bibitem{ref9}
I.~{Grzegorczyk}, M.~{Soliński}, M.~{Łepek}, A.~{Perka}, J.~{Rosiński},
  J.~{Rymko}, K.~{Stępień}, and J.~{Gierałtowski}, ``Pcg classification
  using a neural network approach,'' in \emph{2016 Computing in Cardiology
  Conference (CinC)}, 2016, pp. 1129--1132.

\bibitem{ref10}
F.~Plesinger, I.~Viscor, J.~Halamek, J.~Jurco, and P.~Jurak, ``Heart sounds
  analysis using probability assessment,'' \emph{Physiological measurement},
  vol.~38, no.~8, p. 1685, 2017.

\bibitem{ref11}
C.~Potes, S.~Parvaneh, A.~Rahman, and B.~Conroy, ``Ensemble of feature-based
  and deep learning-based classifiers for detection of abnormal heart sounds,''
  in \emph{2016 Computing in Cardiology Conference (CinC)}.\hskip 1em plus
  0.5em minus 0.4em\relax IEEE, 2016, pp. 621--624.

\bibitem{ref12}
X.~J. Zhu, ``Semi-supervised learning literature survey,'' University of
  Wisconsin-Madison Department of Computer Sciences, Tech. Rep., 2005.

\bibitem{ref13}
B.~Settles, ``Active learning literature survey,'' University of
  Wisconsin-Madison Department of Computer Sciences, Tech. Rep., 2009.

\bibitem{ref14}
D.~Chamberlain, R.~Kodgule, D.~Ganelin, V.~Miglani, and R.~R. Fletcher,
  ``Application of semi-supervised deep learning to lung sound analysis,'' in
  \emph{2016 38th Annual International Conference of the IEEE Engineering in
  Medicine and Biology Society (EMBC)}.\hskip 1em plus 0.5em minus 0.4em\relax
  IEEE, 2016, pp. 804--807.

\bibitem{ref15}
A.~I. Humayun, M.~Khan, S.~Ghaffarzadegan, Z.~Feng, T.~Hasan \emph{et~al.},
  ``An ensemble of transfer, semi-supervised and supervised learning methods
  for pathological heart sound classification,'' \emph{arXiv preprint
  arXiv:1806.06506}, 2018.

\bibitem{ref16}
A.~Ukil, S.~Bandyopadhyay, C.~Puri, R.~Singh, and A.~Pal, ``Class augmented
  semi-supervised learning for practical clinical analytics on physiological
  signals,'' \emph{arXiv preprint arXiv:1812.07498}, 2018.

\bibitem{ref17}
M.~M. Rahman and D.~Davis, ``Addressing the class imbalance problem in medical
  datasets,'' \emph{International Journal of Machine Learning and Computing},
  vol.~3, no.~2, p. 224, 2013.

\bibitem{ref18}
G.~Amit, N.~Gavriely, and N.~Intrator, ``Cluster analysis and classification of
  heart sounds,'' \emph{Biomedical Signal Processing and Control}, vol.~4,
  no.~1, pp. 26--36, 2009.

\bibitem{ref43}
M.~A. Pimentel, D.~A. Clifton, L.~Clifton, and L.~Tarassenko, ``A review of
  novelty detection,'' \emph{Signal Processing}, vol.~99, pp. 215--249, 2014.

\bibitem{ref19}
D.~Ellis, ``Chroma feature analysis and synthesis,'' \emph{Resources of
  Laboratory for the Recognition and Organization of Speech and Audio-LabROSA},
  2007.

\bibitem{ref20}
D.-N. Jiang, L.~Lu, H.-J. Zhang, J.-H. Tao, and L.-H. Cai, ``Music type
  classification by spectral contrast feature,'' in \emph{Proceedings. IEEE
  International Conference on Multimedia and Expo}, vol.~1.\hskip 1em plus
  0.5em minus 0.4em\relax IEEE, 2002, pp. 113--116.

\bibitem{ref21}
C.~Harte, M.~Sandler, and M.~Gasser, ``Detecting harmonic change in musical
  audio,'' in \emph{Proceedings of the 1st ACM workshop on Audio and music
  computing multimedia}.\hskip 1em plus 0.5em minus 0.4em\relax ACM, 2006, pp.
  21--26.

\bibitem{ref22}
N.~V. Chawla, K.~W. Bowyer, L.~O. Hall, and W.~P. Kegelmeyer, ``Smote:
  synthetic minority over-sampling technique,'' \emph{Journal of artificial
  intelligence research}, vol.~16, pp. 321--357, 2002.

\bibitem{ref23}
M.~Oquab, L.~Bottou, I.~Laptev, and J.~Sivic, ``Learning and transferring
  mid-level image representations using convolutional neural networks,'' in
  \emph{Proceedings of the IEEE conference on computer vision and pattern
  recognition}, 2014, pp. 1717--1724.

\bibitem{ref24}
K.~He, X.~Zhang, S.~Ren, and J.~Sun, ``Deep residual learning for image
  recognition,'' in \emph{Proceedings of the IEEE conference on computer vision
  and pattern recognition}, 2016, pp. 770--778.

\bibitem{ref25}
C.~Szegedy, V.~Vanhoucke, S.~Ioffe, J.~Shlens, and Z.~Wojna, ``Rethinking the
  inception architecture for computer vision,'' in \emph{Proceedings of the
  IEEE conference on computer vision and pattern recognition}, 2016, pp.
  2818--2826.

\bibitem{ref26}
G.~Huang, Z.~Liu, L.~Van Der~Maaten, and K.~Q. Weinberger, ``Densely connected
  convolutional networks,'' in \emph{Proceedings of the IEEE conference on
  computer vision and pattern recognition}, 2017, pp. 4700--4708.

\bibitem{ref27}
K.~Simonyan and A.~Zisserman, ``Very deep convolutional networks for
  large-scale image recognition,'' \emph{arXiv preprint arXiv:1409.1556}, 2014.

\bibitem{ref37}
V.~Nair and G.~E. Hinton, ``Rectified linear units improve restricted boltzmann
  machines,'' in \emph{Proceedings of the 27th international conference on
  machine learning (ICML-10)}, 2010, pp. 807--814.

\bibitem{ref36}
N.~Srivastava, G.~Hinton, A.~Krizhevsky, I.~Sutskever, and R.~Salakhutdinov,
  ``Dropout: a simple way to prevent neural networks from overfitting,''
  \emph{The Journal of Machine Learning Research}, vol.~15, no.~1, pp.
  1929--1958, 2014.

\bibitem{ref28}
S.~Hochreiter and J.~Schmidhuber, ``Long short-term memory,'' \emph{Neural
  computation}, vol.~9, no.~8, pp. 1735--1780, 1997.

\bibitem{ref32}
I.~Goodfellow, J.~Pouget-Abadie, M.~Mirza, B.~Xu, D.~Warde-Farley, S.~Ozair,
  A.~Courville, and Y.~Bengio, ``Generative adversarial nets,'' in
  \emph{Advances in neural information processing systems}, 2014, pp.
  2672--2680.

\bibitem{ref33}
O.~Chapelle, B.~Scholkopf, and A.~Zien, ``Semi-supervised learning (chapelle,
  o. et al., eds.; 2006)[book reviews],'' \emph{IEEE Transactions on Neural
  Networks}, vol.~20, no.~3, pp. 542--542, 2009.

\bibitem{ref29}
T.~Salimans, I.~Goodfellow, W.~Zaremba, V.~Cheung, A.~Radford, and X.~Chen,
  ``Improved techniques for training gans,'' in \emph{Advances in neural
  information processing systems}, 2016, pp. 2234--2242.

\bibitem{ref31}
L.~M. Manevitz and M.~Yousef, ``One-class svms for document classification,''
  \emph{Journal of machine Learning research}, vol.~2, no. Dec, pp. 139--154,
  2001.

\bibitem{ref30}
F.~T. Liu, K.~M. Ting, and Z.-H. Zhou, ``Isolation forest,'' in \emph{2008
  Eighth IEEE International Conference on Data Mining}.\hskip 1em plus 0.5em
  minus 0.4em\relax IEEE, 2008, pp. 413--422.

\bibitem{ref34}
E.~J. Benjamin, M.~J. Blaha, S.~E. Chiuve, M.~Cushman, S.~R. Das, R.~Deo,
  J.~Floyd, M.~Fornage, C.~Gillespie, C.~Isasi \emph{et~al.}, ``Heart disease
  and stroke statistics-2017 update: a report from the american heart
  association.'' \emph{Circulation}, vol. 135, no.~10, pp. e146--e603, 2017.

\bibitem{ref44}
G.~D. Clifford, C.~Liu, B.~Moody, J.~Millet, S.~Schmidt, Q.~Li, I.~Silva, and
  R.~G. Mark, ``Recent advances in heart sound analysis,'' \emph{Physiological
  measurement}, vol.~38, no.~8, pp. E10--E25, 2017.

\bibitem{ref38}
M.~Zabihi, A.~B. Rad, S.~Kiranyaz, M.~Gabbouj, and A.~K. Katsaggelos, ``Heart
  sound anomaly and quality detection using ensemble of neural networks without
  segmentation,'' in \emph{2016 Computing in Cardiology Conference
  (CinC)}.\hskip 1em plus 0.5em minus 0.4em\relax IEEE, 2016, pp. 613--616.

\bibitem{ref39}
E.~Kay and A.~Agarwal, ``Dropconnected neural network trained with diverse
  features for classifying heart sounds,'' in \emph{2016 Computing in
  Cardiology Conference (CinC)}.\hskip 1em plus 0.5em minus 0.4em\relax IEEE,
  2016, pp. 617--620.

\bibitem{ref40}
I.~J.~D. Bobillo, ``A tensor approach to heart sound classification,'' in
  \emph{2016 Computing in Cardiology Conference (CinC)}.\hskip 1em plus 0.5em
  minus 0.4em\relax IEEE, 2016, pp. 629--632.

\bibitem{ref41}
J.~Rubin, R.~Abreu, A.~Ganguli, S.~Nelaturi, I.~Matei, and K.~Sricharan,
  ``Classifying heart sound recordings using deep convolutional neural networks
  and mel-frequency cepstral coefficients,'' in \emph{2016 Computing in
  Cardiology Conference (CinC)}.\hskip 1em plus 0.5em minus 0.4em\relax IEEE,
  2016, pp. 813--816.

\bibitem{ref42}
X.~Yang, F.~Yang, L.~Gobeawan, S.~Y. Yeo, S.~Leng, L.~Zhong, and Y.~Su, ``A
  multi-modal classifier for heart sound recordings,'' in \emph{2016 Computing
  in Cardiology Conference (CinC)}.\hskip 1em plus 0.5em minus 0.4em\relax
  IEEE, 2016, pp. 1165--1168.

\bibitem{ref35}
H.~Zhang, M.~Cisse, Y.~N. Dauphin, and D.~Lopez-Paz, ``mixup: Beyond empirical
  risk minimization,'' \emph{arXiv preprint arXiv:1710.09412}, 2017.

\end{thebibliography}

\end{document}